# URBAN OZONE CONCENTRATION FORECASTING WITH ARTIFICIAL NEURAL NETWORK IN CORSICA

**WANI TAMAS** - PhD student, University of Corsica, Laboratory Sciences for Environment UMR CNRS 6134, Ajaccio, France, e-mail: tamas@univ-corse.fr
**GILLES NOTTON** - Assistant Professor, PhD, University of Corsica, Laboratory Sciences for Environment UMR CNRS 6134, Ajaccio, France, e-mail: gilles.notton@univ-corse.fr
**CHRISTOPHE PAOLI** - Assistant Professor, PhD, University of Corsica, Laboratory Sciences for Environment UMR CNRS 6134, Ajaccio, France, e-mail: christophe.paoli@univ-corse.fr
**CYRIL VOYANT** - Researcher, PhD, CHD Castelluccio, radiophysics unit, B.P85 20177 Ajaccio, France, e-mail: voyant@univ-corse.fr
**MARIE-LAURE NIVET** - Assistant Professor, PhD, University of Corsica, Laboratory Sciences for Environment UMR CNRS 6134, Ajaccio, France, e-mail: nivet@univ-corse.fr
**AURELIA BALU** - MSc Student, University of Corsica, Laboratory Sciences for Environement UMR CNRS 6134, Ajaccio, France, e-mail: balu@univ-corse.fr

**Abstract:** Atmospheric pollutants concentration forecasting is an important issue in air quality monitoring. Qualitair Corse, the organization responsible for monitoring air quality in Corsica (France), needs to develop a short-term prediction model to lead its mission of information towards the public. Various deterministic models exist for local forecasting, but need important computing resources, a good knowledge of atmospheric processes and can be inaccurate because of local climatical or geographical particularities, as observed in Corsica, a mountainous island located in the Mediterranean Sea. As a result, we focus in this study on statistical models, and particularly Artificial Neural Networks (ANNs) that have shown good results in the prediction of ozone concentration one hour ahead with data measured locally. The purpose of this study is to build a predictor realizing predictions of ozone 24 hours ahead in Corsica in order to be able to anticipate pollution peaks formation and to take appropriate preventive measures. Specific meteorological conditions are known to lead to particular pollution event in Corsica (e.g. Saharan dust events). Therefore, an ANN model will be used with pollutant and meteorological data for operational forecasting. Index of agreement of this model was calculated with a one year test dataset and reached 0.88.

## 1. Introduction

Tropospheric ozone is a major air pollution problem, both for public health and for environment. Ozone is not directly emitted by human activities. In troposphere, it is a secondary pollutant which formation depends on a complex cycle [1,2]. Ozone is produced by atmospheric photochemical reactions that need solar radiation. Its production is lead by volatile organic compounds and nitrogen oxides concentrations, both emitted by anthropogenic activities. Ozone concentration trend is increasing due to the growth of emissions of its precursors, and many countries are now equipped with an air quality monitoring network which follows the ozone concentration at the ground-levels.

In France, this monitoring is performed by regional Air Quality Monitoring Agreed Associations (AQMAA), thereby allowing the state to take appropriate measures to ensure a good air quality. Air quality forecasting is an important tool that allows authorities to properly react in view to limit anthropogenic pollutants emissions when a pollution peak is predicted. AQMAAs use different forecasting models, according to the characteristics of their regions. Air quality forecasting models have been reviewed recently [3,4]. In this review deterministic models are distinguished from statistical models. The principle of deterministic models is to solve differential equations that describe atmospheric state. Those models are used in more general conditions than statistical models but are more complex and demand important computational resources as well as a good knowledge of atmospheric processes and pollutant sources. Statistical models need local data of variables in relationship with the predicted variable. They are difficult to interpret but can outperform determinist models. Such models can be a good solution to develop a forecasting tool if pollution observed data are available.

This work presents a data-based forecasting model for the French island of Corsica. This island is seated in Mediterranean Sea, in the south of France and west of Italy (Fig. 1). Corsica has an alpine geography, with its highest mountain culminating at 2706m, and its average altitude of 568m. The island has a Mediterranean climate and is exposed to winds such as Sirocco or Mistral, which confer on the island a complex meteorology and can bring pollution plumes from Italy, France or Africa. This particular situation adds difficulties to determinist air quality forecasting. Qualitair Corse is the AQMAA in charge of air quality monitoring in Corsica Island.

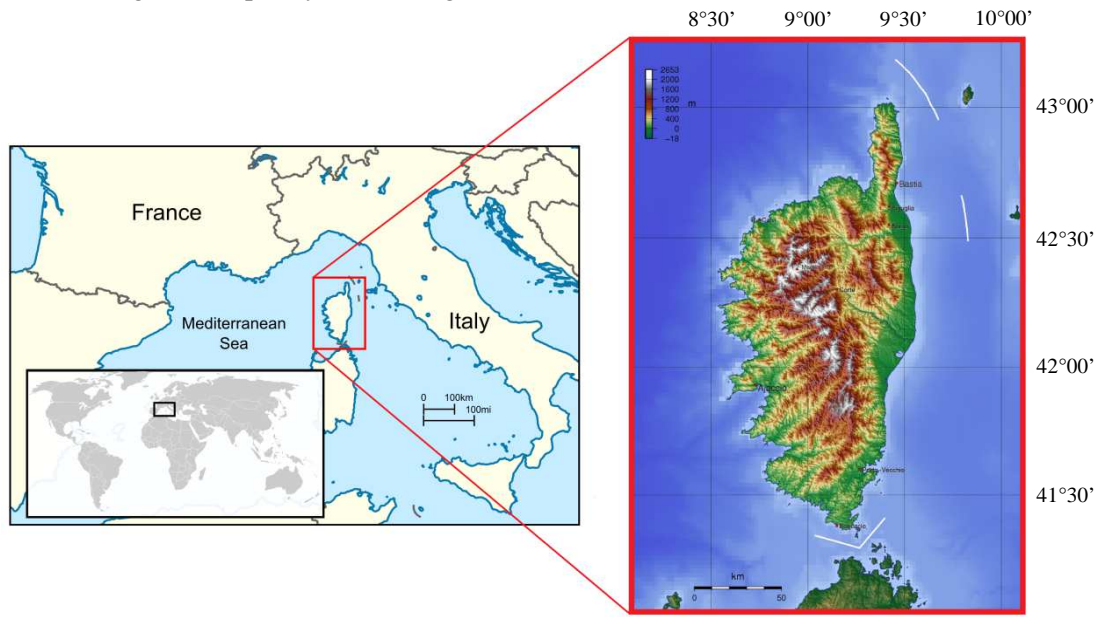

**Fig. 1 -** Position of the Corsica island in the occidental Mediterranean sea and Corsica map showing the mountainous part of the island

In this study we use pollutant data recorded by Qualitair Corse and solar radiation data from Météo France, the French national weather service. We built a new predicting model based on artificial neural networks trained with those data (see next section for details), for hourly ozone concentration predictions 24 hours ahead. We will first introduce our model in section 2 before presenting the data operated by the model in section 3. Our experiments are shown in section 4 and their results discussed in section 5. The last part presents the conclusions of this study and some future perspectives.

## 2. Models

Artificial Neural Networks (ANNs) are statistical models of artificial intelligence family, able to learn complex relationships between inputs and outputs. They were developed so as to model how the human brain processes information in the 40s [5].

ANNs are numerical calculators whose structure is inspired by biological neurons model. An ANN has a parallel-distributed structure and consists of a set of processing elements called neurons or nodes. The ANN structure is composed of an input layer which receives data, an output layer to send computed information, and one or several hidden layers linking the input and output layers.

Applications of ANNs in atmospheric sciences were reviewed in the late 90s [6]. They have often been used as time series forecasting models, and show good results in this domain [7]. There are various types of ANN architectures which fit for different modeling patterns. According to the chosen architecture, all or a part of the neurons in a layer are connected with all or a part of the neurons of the previous and next layer. The number of hidden layers and of neurons in each layer depends on the specific model, convergence speed, generalization capability, physical processes and training data that the network will simulate [8].

Among possible architectures, Multilayer Perceptron (MLP) is distinguished by its universal approximator capabilities [9], it can find any smooth mesurable relationship between predictands and predictors variables. This structure was used to predict daily maximum ozone concentration and compared to other statistical methods with better performances [8]. MLP was also used in the past decade to forecast hourly ozone level for a 8h horizon [10,11]. More recently, some works have shown that this neural network architecture could be used to make accurate hourly prediction one day ahead [12,13]. Preliminary work with ANN in Corsica already led to a model for 1 hour ahead ozone concentration prediction [14].

A formal neuron, elementary constituant of an ANN, transforms the input variables it receives in an output variable. Each input $x_i$ of a neuron is multiplicated by a specific weight $w_i$. A bias $b$ is added to the sum of all products $x_i w_i$ and the result becomes the argument of the neuron's activation function $f$ which gives the output variable. The mathematical equivalent of a neuron with $N$ inputs is a function of the type :

$$y = f\left(\sum_{i=1}^{N} w_i \cdot x_i + b\right) \quad (1)$$

MLP is a feed-forward neural network constitued by several layers of neurons, each one receiving inputs from the previous layer and communicating their outputs to the next layer (see Fig. 2). The input layer consists of the input data, it is connected to an internal layer of neurons called hidden layer. There may be a variable number of these interconected hidden layers. The last layer of neurons is called output layer and produces the outputs of the network. It is important to note that the input layer is not composed by neurons but by input data.

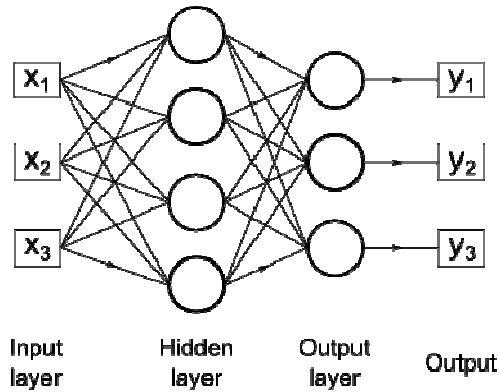

**Fig. 2** - Shematic view of a MLP with 3 inputs, 4 hidden nodes and 3 outputs

The value of weigths and biases of a MLP is determined during a supervised learning phase. Purpose of this learning phase is to minimise the mean squared error between the output of the MLP and a target dataset by addapting network's weigths and biases. During the learning step, input and target training data sets are provided to the network. The training algorithm searches the best weights/biases configuration to capture the underlying relationship between input and target data.

The forecasting nature of the model lies in the choice of input and target data set for the training phase. Those data sets form time series. The target set is created by shifting forward in time the forecasted time series. During the training, the network will try to find the best configuration to reproduce relationships between input data and future value of target data. Mathematical equivalent of a trained MLP is a simple non-linear regression of inputs, shown in Eq. 2 for a network with one hidden layer.

$$y = g\left(\sum_{j=1}^{Nh}\left(w_{jo} \cdot \left(f(\sum_{i=1}^{Ni} w_{ij} \cdot x_i + b_j)\right) + b_o\right)\right) \quad (2)$$

with $i$ the index of inputs, and $j$ the index of neurons in the hidden layer,
$N_i$ the number of inputs and $N_h$ the number of neurons in the hidden layer,
$x_i$ the inputs and $y$ the output,
$f$ and $g$ respectively the activation functions of hidden neurons and of output neuron,
$w_{ij}$ the weight between the input $i$ and the hidden neuron $j$ and $b_j$ the bias in hidden neuron $j$,
$w_{jo}$ the weight between hidden neuron $j$ and the output neuron and $b_o$ the bias in the output neuron.

To evaluate the performances of this model, we built a simple persistence model, whose principle is to retain the actual value $x$ of a time series as a forecasted value $\bar{x}$ (see Eq. 3).

$$\bar{x}(t + 24) = x(t) \qquad (3)$$

It takes sense since we are making predictions 24 hours ahead and the ozone concentration time series has a periodical component of 24 hours. All computations were executed using Matlab $^{©}$.

## 3. Data

The air pollution data used in this study was measured and collected by Qualitair Corse, the AQMAA responsible for air pollution monitoring in Corsica. The association has nine fixed automatic monitoring stations which are distributed over the island (Table 1 and Fig. 3). Corsica is a mountainous island and most of the population concentrates on the coasts. Ajaccio and Bastia are the two biggest cities, and the two main population basins. For public health purposes, air quality monitoring focuses on those two regions that gather both the majority of the population and the main emission sources of the island.

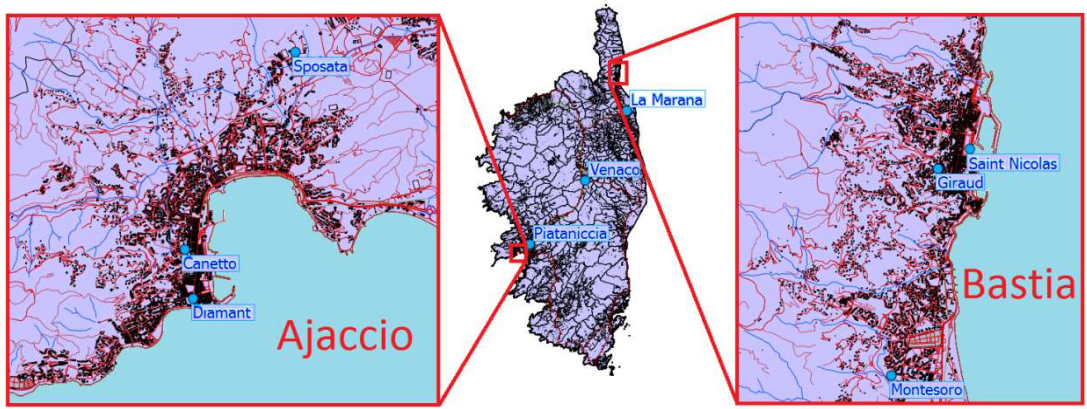

**Fig. 3** - Monitoring stations in Corsica (Base map provided by IGN)

The stations are classified into five categories: traffic, urban, suburban, industrial and rural measuring several atmospheric pollutants ($NO_2$, Ozone, $SO_2$, small (PM2.5) and large (PM10) particles). We have one station of the four first types in Ajaccio and Bastia and a rural one in Venaco (centre of Corsica).

Table 1

**Measuring stations in Corsica**

| Area | Stations | Category | Height above mean sea level (m) | Measured pollutants |
|---|---|---|---|---|
| Bastia | Giraud | Urban | 60 | $NO_2$ $O_3$ PM10 $SO_2$ |
| | Montesoro | Suburban | 47 | $NO_2$ $O_3$ PM2.5 |
| | St Nicolas | Traffic | 5 | $NO_2$ PM10 |
| | La Marana | Industrial | 15 | $NO_2$ $O_3$ PM10 $SO_2$ |
| Venaco | Venaco | Rural | 653 | $NO_2$ $O_3$ PM10 PM2.5 |
| Ajaccio | Canetto | Urban | 39 | $NO_2$ $O_3$ PM10 $SO_2$ |
| | Sposata | Suburban | 60 | $NO_2$ $O_3$ |
| | Piataniccia | Industrial | 30 | $NO_2$ $O_3$ PM10 $SO_2$ |
| | Diamant | Traffic | 12 | $NO_2$ PM10 |

In this paper, we only used pollutant data recorded at the two urban and the two suburban stations: Canetto and Sposata in Ajaccio and Giraud and Montesoro in Bastia. Those stations are representative of population exposure to air pollutants in the two cities. Five years of data were available for this study, between the beginning of 2008 and the end of 2012. Global solar radiation data were provided by Météo France. Other meteorological variables were recorded at Sposata station and Montesoro station.

All data were hourly averages (measures are done every 10 seconds, automatic stations send averaged data every 15 minutes and then hourly averages are calculated). Endogenous and exogenous data were used as inputs of the network. Exogenous data were pollution and weather data. Because of its important role in the ozone production cycle [1], nitrogen dioxide concentration ($NO_2$) measured by the station was used. Meteorological variables (wind force, wind direction, global solar radiation, temperature and precipitation, called below MET dataset) are known to influence ozone concentration [2] and were included into the exogenous data set.

We use $O_3$ concentration hourly time series shifted of 24 time steps as target dataset, representing $O_3$ concentration at *h+24*. When data are inputted into the network in view to make an hourly prediction at *h+24*, several values of each observed predictor at different time lags are passed to the hidden layer. Those lags were chosen using Average Mutual Information (AMI, see Eq. 4). Mutual information (measured in bits) is a quantitative measure of statistical dependence between two dataset.

$$AMI(x, y) = \sum_{i=1}^{n} \sum_{j=1}^{m} P(x_i, y_j) \log \frac{P(x_i, y_j)}{P(x_i) P(y_j)} \qquad (4)$$

with *x* and *y* two time series,
*n* and *m* the numbers of class to compute time series distributions,
*i* and *j* indexes of classes,
$P(x_i)$ the probability to have *x* value within class *i*,
$P(y_j)$ the probability to have *y* value within class *j*,
$P(x_i, y_j)$ the probability to have *x* value within class *i* and *y* value within class *j*.

We calculated AMI between $O_3$ concentration time series 24 hours in advance and input data time series for every lag between 0 and 72 hours in order to find with which lag the predictor time series shared the most information with $O_3$ concentration at *h+24*. Fig 4 shows AMI of $O_3$ concentration and wind direction. Periodical peaks show the non-stationarity of time series. Although MLP is a stationary estimator, it is suited for non-stationary cases on short horizon.

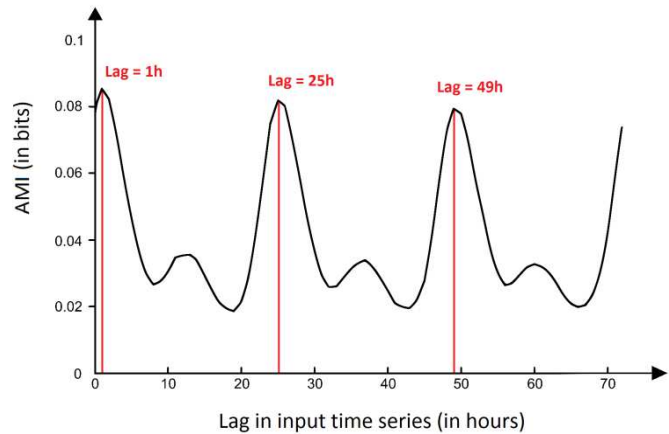

**Fig 4** - AMI of wind direction and ozone concentration at h+24, with the tree lags retained to be passed as inputs

The weekdays bring information related to human activities that impact the air quality (working or non-working days). As such temporal variables have been shown to improve air quality forecasting with ANNs [15], we chose to use weekday number (1 for Monday to 7 for Sunday) in this work. As proposed in several studies [6,12], we also used periodical variables representing the cycle of passing hours: $\sin(2\pi h/24)$ et $\cos(2\pi h/24)$. Those three time indices are below referred as TI dataset.

Both endogenous and exogenous data may present several missing values due to measure interruptions. Those missing values are problematic; they cannot be processed by the model, which mathematical operations are defined for real values. All missing were replaced by the mean value of the variable at the same day of the year and same hour. Other methods for dealing with missing values have been reviewed [16], and will be investigated in further work.

Before computation, all inputs were normalized between -1 and 1. This measure avoids overrepresentation of one predictor in the prediction because of its values range.

## 4. Experiments

We built an ANN model based on the MLP, using endogenous and exogenous data. Levenberg-Marquardt algorithm was used to train the network. Input data were delayed of 24 hours, to train the network to a 24 hours ahead forecasting model.

There are various parameters to settle when creating a MLP, the number of layers, the number of neurons of each layer, the activation function of neurons. With one single hidden layer, a MLP is able to model non-linear relationships between inputs and outputs. We used a one hidden layer MLP, since we observed that adding a second hidden layer did not increased performance. We used various numbers of neurons in the hidden layer and retained the configuration with the best results: 12 hidden neurons. The output layer was composed of one single neuron for one single output. Activation function is a hyperbolic tangent function for the hidden layer and a linear function for the neuron of the output layer.

It is important to avoid over-fitting the training data, to ensure the network generalisation capacity. For that reason, we used the early-stopping method: at each iteration of the training phase, performances of the model are tested on an independent set of data, the validation set. When this performance stops increasing during 6 iterations in a row, the learning phase is interrupted although training performance score keeps growing. Trained MLP is then evaluated with another independent data set, the test set, in order to quantify prediction error. Three years of data composed the training set and validation and test sets were each composed of one year of data.

Performance was evaluated during the training phase with the Mean Squared Error (MSE). During the test phase we calculated Root Mean Squared Error (RMSE), normalised Root Mean Squared Error (nRMSE), Mean Absolute Error (MAE) and Index of Agreement (IA) which are reported below (Eqs. 5 to 6). IA was introduced as an error indices suitable for forecasting model evaluation [17]. It ranges between 0 (worst) and 1 (best) and represents the degree to which prediction is error-free.

$$MSE = \frac{1}{N}\sum_{i=1}^{N}(o_i - p_i)^2 \quad RMSE = \sqrt{\frac{1}{N}\sum_{i=1}^{N}(o_i - p_i)^2} \quad nRMSE = \frac{\sqrt{\frac{1}{N}\sum_{i=1}^{N}(o_i - p_i)^2}}{\bar{o}} \quad (5)$$

$$MAE = \frac{1}{N}\sum_{i=1}^{N}|o_i - p_i| \qquad IA = 1 - \frac{\sum_{i=1}^{N}(p_i - o_i)^2}{\sum_{i=1}^{N}(|p_i - \bar{o}| + |o_i - \bar{o}|)^2} \quad (6)$$

with $N$ the number of samples and $i$ their index,
$o_i$ and $p_i$ observed and predicted concentrations,
$\bar{o}$ the global average of observed concentrations.

We investigated four different input data configurations for data recorded in Canetto, Sposata, Montesoro and Giraud station using successively:
-endogenous data ($O_3$ concentration),
-endogenous and exogenous pollution data ($O_3+NO_2$),
-endogenous and exogenous pollution and meteorological data ($O_3+NO_2+MET$),
-endogenous and exogenous pollution and time indices ($O_3+NO_2+TI$),
-endogenous and exogenous data and time indices ($O_3+NO_2+MET+TI$)

at lags determined using AMI (see Eq. 4 in previous section).

## 5. Results

Results of persistence and MLP models with different inputs are shown in Table 2. Each experiment with MLP (network training + test) was run seven times and we gave the average values of error indices, to study the relationship between predictors and performances. Those results are globally close to other works on $O_3$ concentration forecasting 24 hours in advance with ANNs [12]. We can see that performances of the model are enhanced by the addition of $NO_2$, meteorological variables and time indices as input. However, this improvement remains quite small (about 1% for nRMSE for all stations).

Table 2

**Error indices**

| Station | Model (with input datasets) | RMSE (µg.m$^{-3}$) | nRMSE (%) | MAE (µg.m$^{-3}$) | IA [1] |
|---|---|---|---|---|---|
| Canetto | Persistence | 21.26 | 36.58 | 16.26 | 0.85 |
| | MLP ($O_3$) | 17.94 | 30.44 | 14.30 | 0.86 |
| | MLP ($O_3+NO_2$) | 17.81 | 30.21 | 14.23 | 0.87 |
| | MLP ($O_3+NO_2$+MET) | 17.41 | 29.54 | 13.85 | 0.87 |
| | MLP ($O_3+NO_2$+TI) | **17.31** | **29.37** | **13.66** | **0.88** |
| | MLP ($O_3+NO_2$+MET+TI) | 17.32 | 29.39 | 13.74 | **0.88** |
| Sposata | Persistence | 20.44 | 34.03 | 15.63 | 0.85 |
| | MLP ($O_3$) | 17.50 | 28.35 | 13.88 | 0.85 |
| | MLP ($O_3+NO_2$) | 17.44 | 28.24 | 13.80 | 0.85 |
| | MLP ($O_3+NO_2$+MET) | 16.99 | 27.52 | 13.40 | 0.86 |
| | MLP ($O_3+NO_2$+TI) | 17.19 | 27.84 | 13.57 | 0.86 |
| | MLP ($O_3+NO_2$+MET+TI) | **16.90** | **27.38** | **13.31** | **0.87** |
| Giraud | Persistence | 19.46 | 25.60 | 14.73 | **0.81** |
| | MLP ($O_3$) | 17.19 | 22.69 | 13.39 | 0.79 |
| | MLP ($O_3+NO_2$) | 16.89 | 22.22 | 13.12 | 0.80 |
| | MLP ($O_3+NO_2$+MET) | 16.78 | 22.14 | 13.01 | 0.80 |
| | MLP ($O_3+NO_2$+TI) | 16.61 | 21.92 | 12.92 | **0.81** |
| | MLP ($O_3+NO_2$+MET+TI) | **16.53** | **21.81** | **12.82** | **0.81** |
| Montesoro | Persistence | 18.51 | 24.23 | 14.04 | **0.84** |
| | MLP ($O_3$) | 16.30 | 21.20 | 12.61 | 0.83 |
| | MLP ($O_3+NO_2$) | 16.30 | 20.92 | 12.41 | **0.84** |
| | MLP ($O_3+NO_2$+MET) | 16.06 | 20.90 | 12.39 | **0.84** |
| | MLP ($O_3+NO_2$+TI) | 15.96 | 20.77 | 12.32 | **0.84** |
| | MLP ($O_3+NO_2$+MET+TI) | **15.90** | **20.69** | **12.30** | **0.84** |

[1] IA is dimensionless. Best results written in bold

RMSE improvement due to addition of meteorological data in input dataset is slightly better for suburban stations (0.24 µg.m$^{-3}$ for Montesoro and 0.45 µg.m$^{-3}$ for Sposata) than for the corresponding urban station (0.10 µg.m$^{-3}$ for Giraud and 0.40 µg.m$^{-3}$ for Canetto). The reason can be that those data are recorded in the suburban stations. Suburban area may also be less influenced by pollutant sources than urban station, and thus relatively more dependants to meteorology.

In term of nRMSE, the mean gain between the persistence model and the best MLP is upper than 5%. MLP models as persistence model perform better with stations from Bastia than from Ajaccio, regardless of the setting. IA follows an opposite scheme. This phenomenon could be the consequence of two different ozone concentration dynamics. Ajaccio and Bastia are two coastal cities and are exposed to sea and land breezes. At night, ozone concentration in Bastia stay high while it drops in Ajaccio as expected in normal conditions. This could be the consequence of two different configurations vis-à-vis nocturnal ozone income from rural areas due to the land breeze. Ozone concentration range is therefore larger in Ajaccio, which could explain a bigger RMSE for all models while the better IA seems to show that the models are more precise than in Bastia where ozone dynamics seems to be more complex.

For this station, the persistence model has even reached the same IA than the MLP, though RMSE of the neural network stays better. Other meteorological variables that could help the network to learn the nocturnal ozone behavior, as atmospheric boundary-layer thickness or other wind data in the region (i.e. output of weather forecasting models) could be helpful to improve the predictions.

## 6. Conclusion

A MLP model was built to forecast $O_3$ concentration 24 hours ahead at two urban sites in Corsica. The neural network was trained with pollution and meteorological data, in addition to temporal variables. It showed good results, its performances were better than those of the persistence model used as reference.

This first work on air quality forecasting at horizon *h+24* with Corsican data helped to prepare the building of an operational forecasting model able to detect pollution event. In further work, we will try to improve our predictions by several ways: we will focus on the data pretreatment: alternative way to treat missing data will be investigated, we will work on predictions with time series made stationary and we will consider input data selection with several methods as genetic algorithms or AMI.

After those results, it seems interesting to use new meteorological variables that could represent specific dynamics of pollution concentration in the two cities. Finally, other pollutant forecasting models, for $NO_2$ or fine particles concentration, will also be investigated.